\documentclass{article} 
\usepackage{iclr2025_conference,times}

\usepackage{amsmath,amsfonts,bm}

\def\eqref#1{equation~\ref{#1}}

\def\1{\bm{1}}

\DeclareMathAlphabet{\mathsfit}{\encodingdefault}{\sfdefault}{m}{sl}
\SetMathAlphabet{\mathsfit}{bold}{\encodingdefault}{\sfdefault}{bx}{n}

\usepackage{hyperref}
\usepackage{url}
\usepackage{xspace}
\usepackage{booktabs}
\usepackage{multirow}
\usepackage{enumitem}
\usepackage{wrapfig}
\usepackage{graphicx}
\usepackage{pifont}
\usepackage{xcolor}

\title{Benchmarking General Mobile Assistants in Challenging Real-World Scenarios}

\author{\textbf{Yiqi Zhu\textsuperscript{1}, Feiyu Gao\textsuperscript{3}, Jiaxing Fan\textsuperscript{3}, Jiahui Zeng\textsuperscript{3}, Minggang Wu\textsuperscript{3}} \\
\textbf{Chenliang Li\textsuperscript{3}, Haiyang Xu\textsuperscript{3}, Peng Li\textsuperscript{2,$\dagger$}, Ming Yan\textsuperscript{3,$\dagger$}, Yang Liu\textsuperscript{1,2,$\dagger$}} \\
\textsuperscript{1}Dept. of Comp. Sci. \& Tech., Institute for AI, Tsinghua University, Beijing, China \\
\textsuperscript{2}Institute for AI Industry Research (AIR), Tsinghua University, Beijing, China \\
\textsuperscript{3}Alibaba Token Hub, Alibaba Group \\
\texttt{zhuyq0218@gmail.com, lipeng@air.tsinghua.edu.cn} \\
\texttt{ym119608@alibaba-inc.com, liuyang2011@tsinghua.edu.cn}
}

\newcommand{\method}{\textsc{GMA}\xspace}
\newcommand{\cmark}{\textcolor{green!60!black}{\ding{51}}}
\newcommand{\xmark}{\textcolor{red!70!black}{\ding{55}}}

\iclrfinalcopy 
\begin{document}

\maketitle

\begin{abstract}
Graphical user interfaces have emerged as a primary environment for evaluating autonomous AI agents on multimodal interactive tasks. Existing benchmarks such as AndroidWorld and MobileWorld have established strong foundations for evaluating mobile agents in emulator-based Android tasks, but they remain limited in two respects. First, their application coverage does not yet fully capture the vast diversity of Android device ecosystems. Second, there is room for their task designs to be extended to encompass the nuanced complexity of realistic user requirements. To further propel the evaluation of mobile agents, we present \method, a benchmark for assessing general mobile assistants in challenging real-world scenarios. To improve application ecosystem coverage, we manually develop seven applications based on open-source projects, spanning practical domains such as lifestyle sharing and travel planning. To better capture real-world user needs, we construct 300 tasks across four difficulty tiers, ranging from simple atomic actions to highly complex multi-step workflows. Together, these applications and tasks provide a more comprehensive setting for evaluating mobile agents under diverse and demanding conditions. We evaluate eight frontier models on \method, and results show that current mobile agents remain far from reliably handling realistic user requirements, even when they perform well on atomic interactions. Beyond model evaluation, we conduct controlled ablation studies on several agent harness choices, including context retention and explicit state tracking, under a shared environment, model setting, and task taxonomy. Our results show that appropriate harness design can meaningfully improve agent performance, with particularly clear benefits on demanding workflows, while the effectiveness of specific designs can also vary across foundation models. Overall, \method complements existing benchmarks by expanding both the breadth of the evaluated application ecosystem and the depth of task complexity. It exposes important limitations of current mobile agents and provides a challenging testbed for both evaluating model capabilities and studying how agent harness design can support reliable execution in complex mobile workflows.

\end{abstract}

\noindent{\let\thefootnote\relax
\footnotetext{$^\dagger$ Corresponding Authors. Code Available at \url{https://github.com/Tongyi-Zhiwen/GMA}.}}

\section{Introduction}

\begin{figure}[t]
\centering
\includegraphics[width=\textwidth]{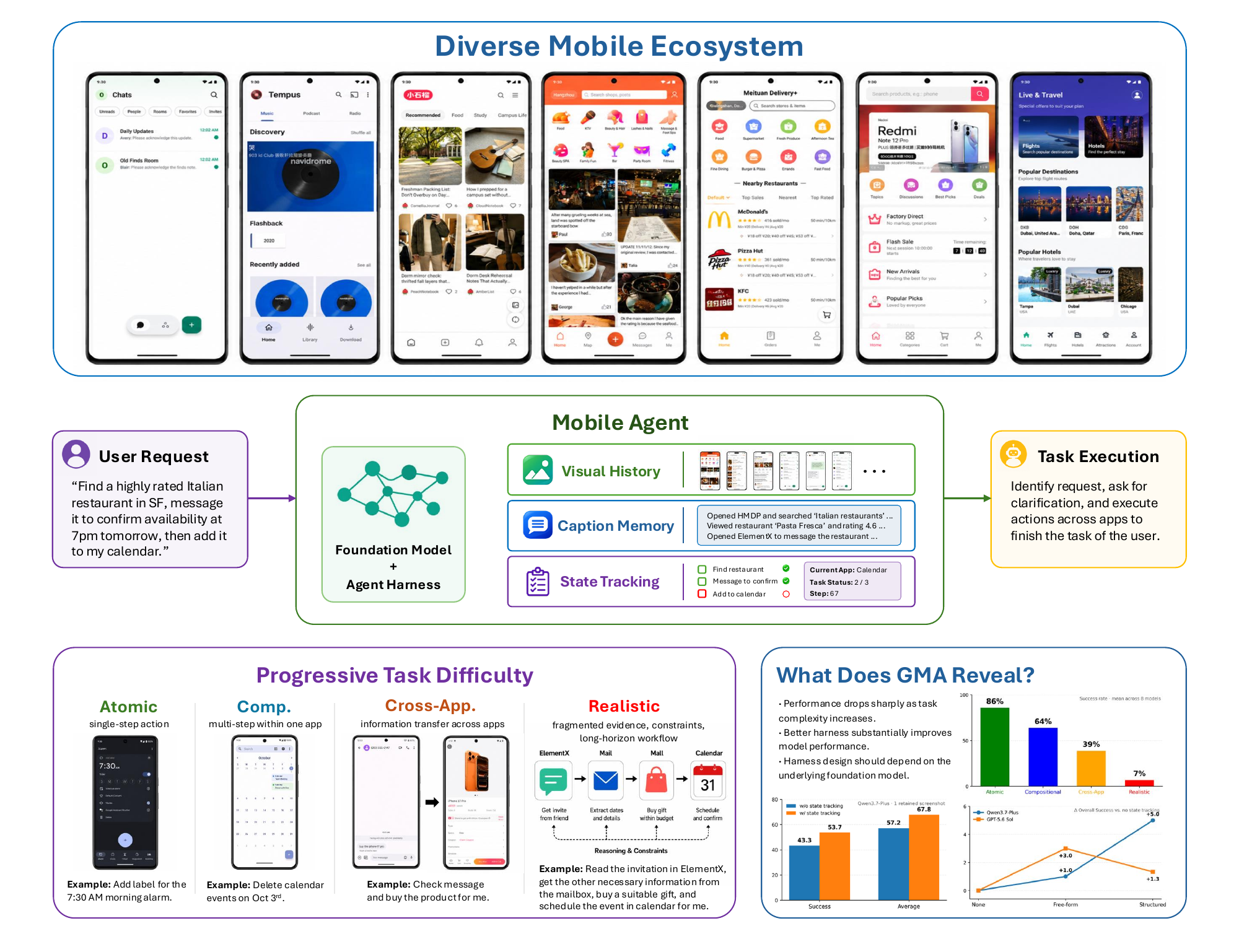}
\vspace{-2em}
\caption{\textbf{Overview of \method}. \method provides a diverse and reproducible mobile ecosystem with seven custom applications, together with 300 tasks spanning four progressively more demanding tiers: Atomic, Compositional, Cross-Application, and Realistic. We further study the role of the agent harness, including context retention, explicit state tracking, and model-specific harness design. Evaluation on \method reveals a substantial performance degradation as task complexity increases and highlights the growing importance of effective harness design for complex mobile workflows.}
\vspace{-1em}
\label{fig:teaser}
\end{figure}

Recent advances in large language models (LLMs) have empowered agents to move beyond static language tasks and interact autonomously with external environments~\citep{wei2025browsecompsimplechallengingbenchmark, merrill2026terminalbenchbenchmarkingagentshard, datacurve2026deepswev11}. Among diverse scenarios, graphical user interfaces (GUIs)~\citep{OSWorld, yuan2026osworld20benchmarkingcomputeruse} and mobile use~\citep{rawles2024androidworlddynamicbenchmarkingenvironment, kong2025mobileworld} serve as an important testbed for evaluating agent capabilities. Successfully navigating GUI tasks requires agents to perceive visual layouts, comprehend element functions, translate high-level user intentions into precise action sequences, and adapt to dynamic application states. Furthermore, realistic mobile workflows typically demand long-horizon planning, information retrieval, constraint satisfaction, and error recovery. Such long interaction trajectories also place substantial demands on how agents preserve historical observations, intermediate information, and task progress throughout execution~\citep{li2025mobileuse, shi2026androtmeminteractiontrajectoriesanchored}. Together, these characteristics make GUI environments an ideal setting for assessing the capability of agents to reliably execute complex, real-world tasks.

Existing benchmarks have established a strong foundation for evaluating mobile agents within executable Android environments. AndroidWorld~\citep{rawles2024androidworlddynamicbenchmarkingenvironment} introduces a reproducible evaluation framework with dynamically instantiated tasks and state-based verification, serving as a canonical benchmark for mobile-agent evaluation. As a step forward, MobileWorld~\citep{kong2025mobileworld} extends AndroidWorld by integrating three additional applications and designing long-horizon, cross-application tasks. It also incorporates user interaction and MCP-augmented tasks to move beyond purely GUI-driven operations. These two benchmarks have substantially advanced the evaluation of mobile agents, and provided essential infrastructure for studying increasingly capable models.

Building on this progress, we explore how mobile agent evaluation can be further expanded along two dimensions. First, we aim to broaden application coverage so that the evaluation environment reflects a wider range of functional domains, interface designs, and interaction patterns characteristic of real Android ecosystems. Second, we seek to expand task complexity by introducing workflows that capture additional characteristics of practical mobile use, which include multi-step requirements, cross-state dependencies, and the integration of information retrieval, reasoning, and execution. These extensions provide a broader setting for uncovering the capabilities and limitations of current mobile agents.

To this end, we present \method, a benchmark for evaluating general mobile assistants in challenging and complex mobile scenarios. To construct a functionally diverse and highly interactive Android environment, we develop seven custom applications based on open-source projects. These applications model common categories of digital services, including instant messaging, music library management, lifestyle sharing, local discovery, on-demand delivery, e-commerce, and travel planning. Together, they substantially broaden the range of mobile functions and interaction patterns available within a reproducible evaluation environment. Within this ecosystem, we meticulously design 300 tasks categorized into four difficulty tiers. This progressive structure evaluates capabilities ranging from atomic operations within a single interface to complex multi-step workflows requiring long-horizon planning, information synthesis, and cross-application reasoning. It allows us to systematically analyze capability bottlenecks as task complexity increases, while also providing a common setting for studying how different agent mechanisms behave across increasingly demanding tasks.

We evaluate several frontier model families, including Qwen, GPT and Claude series, on \method. Our empirical analysis reveals that even models demonstrating strong proficiency on simpler tasks still struggle significantly on the more demanding workflows in our environment, and agent performance consistently degrades as task complexity increases. While the strongest models achieve high success rates on atomic interactions, no evaluated model successfully completes more than 20\% of the realistic tasks, and the highest Average score on this tier is only 45.27. These results highlight the substantial difficulty of reliably executing intricate, real-world mobile workflows.

We further conduct controlled ablation studies to examine how agent harness design influences mobile task execution. Although previous GUI agents have incorporated mechanisms such as planning, reflection, and memory, their effects are often evaluated as part of complete systems in which multiple components vary simultaneously. We instead isolate several context management and state tracking choices under a shared environment and task taxonomy. Our results show that retaining richer historical information, including previous visual states and compact captions of discarded observations, can improve agent performance, with particularly clear benefits on demanding workflows. Explicitly maintaining task requirements and execution progress can also provide additional gains especially for realistic tasks. We further observe that the same state tracking design can affect different foundation models differently, suggesting that harness mechanisms may not transfer uniformly across models. Overall, these findings indicate that performance on complex mobile tasks depends not only on the capabilities of the underlying foundation model, but can also be meaningfully influenced by how the surrounding agent harness manages historical context and execution state, motivating further study of harness design alongside continued advances in foundation models.

\section{Related Work}

\subsection{GUI and Mobile Evaluation}

Agent evaluation has evolved from static prediction tasks toward interactive execution in dynamic environments. In the web domain, WebShop~\citep{NEURIPS2022_82ad13ec} and WebArena~\citep{zhou2023webarena} evaluate whether agents can navigate websites and complete user-specified tasks through sequences of grounded actions. In software engineering, SWE-bench~\citep{jimenez2024swebench} and SWE-Bench Pro~\citep{deng2025swebenchproaiagents} assess agents on resolving repository-level issues, while ProgramBench~\citep{yang2026programbenchlanguagemodelsrebuild} evaluates whether agents can reconstruct complete programs from their documentation and observable behavior. BrowseComp~\citep{wei2025browsecompsimplechallengingbenchmark} and DeepResearch Bench~\citep{du2025deepresearch} instead focus on information-seeking agents, testing persistent web search, evidence collection, and the synthesis of research reports. More recently, Workspace-Bench~\citep{tang2026workspacebench10benchmarkingai} and AutomationBench~\citep{shepard2026automationbench} place agents in realistic workspace and business-process environments, requiring them to reason over heterogeneous files or coordinate operations across multiple software services. Together, these benchmarks illustrate the increasingly broad scope of agent evaluation, spanning web navigation, software engineering, information research, and workplace automation.

Among these settings, graphical user interfaces have emerged as a particularly important scenario because they require agents to jointly perceive visual content, ground language instructions to interface elements, plan action sequences, and adapt to changing application states. Early GUI evaluation primarily focused on isolated perception and action prediction capabilities in static or offline settings. ScreenSpot~\citep{cheng2024seeclick}, for example, measures whether a model can locate the interface element corresponding to a textual instruction from a single screenshot. Android in the Wild~\citep{rawles2023androidwildlargescaledataset} evaluates action prediction over recorded mobile interaction trajectories. Although these benchmarks provide important diagnostics of visual grounding and interface understanding, they do not fully capture closed-loop interaction, where each action changes the subsequent environment state and agents must respond to execution errors and unexpected outcomes. More recent agentic benchmarks therefore evaluate end-to-end task completion in executable GUI environments. OSWorld~\citep{OSWorld} provides a unified environment for agents to interact with real desktop applications and operating system interfaces, using task-specific initialization and execution-based verification. AndroidWorld~\citep{rawles2024androidworlddynamicbenchmarkingenvironment} establishes a corresponding evaluation framework for mobile agents, featuring dynamically instantiated tasks, direct interaction with an Android emulator, and programmatic state-based verification. These two benchmarks have become canonical benchmarks for evaluating modern agents in desktop and mobile GUI environments.

Recent mobile benchmarks have extended interactive evaluation along complementary dimensions. MobileWorld~\citep{kong2025mobileworld} introduces longer cross-application workflows, agent-user interaction, and MCP-augmented tasks within a reproducible Android environment. MobileBench-OL~\citep{wu2026mobilebench} evaluates online execution, reasoning, exploration, and robustness to environmental noise across a large collection of Chinese applications. PSPA-Bench~\citep{nie2026pspabenchpersonalizedbenchmarksmartphone} focuses on personalized mobile assistance through user-specific instructions and fine-grained process evaluation. AndroidDaily~\citep{sui2026androiddailyverifiablebenchmarkmobile} evaluates agents on closed-source commercial applications using observable interaction trajectories to assess task completion without access to internal application states. MobileGym~\citep{wu2026mobilegymverifiablehighlyparallel} develops a lightweight, browser-hosted mobile simulation platform with several web applications that simulate real mobile apps, supporting deterministic state-based evaluation and highly parallel rollouts for agent training. iOSWorld~\citep{jang2026iosworldbenchmarkpersonallyintelligent} extends interactive mobile evaluation beyond Android by introducing a native iOS environment with persistent user data and tasks requiring personalization and cross-application reasoning. Together, these benchmarks broaden mobile-agent evaluation across application coverage, personalization, platform diversity, environmental realism, and training scalability.

\subsection{Agent Harness}

The concept of agent harness has recently been used to describe the system surrounding a foundation model that enables and structures its interaction with tools and environments~\citep{anthropic2025buildingagents, anthropic2025effectiveharnesses}. The underlying idea, actually, is not new in agent research, as many earlier systems had already incorporated mechanisms that organize the interaction loop, manage context, and connect models with external actions. ReAct~\citep{yao2022react} interleaves reasoning, actions, and environment observations, providing a general framework task solving agents. Reflexion~\citep{shinn2023reflexionlanguageagentsverbal} further augments agents with linguistic feedback, self-reflection, and episodic memory to improve behavior across repeated attempts. In software engineering, SWE-agent~\citep{yang2024sweagent} introduces an agent-computer interface for repository navigation, file editing, and test execution. OpenHands~\citep{wang2025openhands} provides a broader platform integrating code editing, terminal execution, web browsing, and sandbox management. Viewed through the contemporary notion of the agent harness, these works demonstrate how external scaffolding can organize model behavior and adapt foundation models to complex interactive tasks.

Following the introduction of this terminology, recent work has increasingly treated the harness as a first-class component that can be deliberately designed, evaluated, and improved. ~\cite{openai2026harnessengineering} demonstrates how repository structure, documentation, testing infrastructure, observability, and feedback loops can make a large software project legible to coding agents and progressively increase their autonomy. ~\cite{lin2026agenticharnessengineeringobservabilitydriven} introduces an observability-driven framework that automatically evolves harness components such as tools, middleware, and long-term memory from execution experience. Harness-Bench~\citep{harnessbench2026} systematically compares configuration-level harness effects under shared tasks and execution conditions, showing that agent performance should not be attributed to the foundation model alone.

For GUI agents, harness plays an important role in coordinating perception, planning, memory, verification, and execution over long interaction trajectories. HiconAgent~\citep{zhou2026hiconagenthistorycontextawarepolicy} improves the use of interaction history through dynamic context sampling and history compression, reducing computational overhead while preserving task-relevant information. Mobile-Agent-v3.5~\citep{xu2026mobile} employs a multi-agent harness that assigns planning, execution, action verification, and persistent memory to specialized Manager, Worker, Reflector, and Notetaker roles. PhoneHarness~\citep{li2026phoneharnessharnessingphoneuseagents} coordinates GUI interactions, device-side commands, and structured tools through a unified execution loop with deterministic action routing and auditable traces. ~\cite{zheng2026taskstaterepresentationlonghorizonmobile} introduces a training-free external wrapper that maintains a global instruction summary, a dynamic subgoal-progress tracker, and a transition-aware action verifier to guide the reasoning process of GUI agents. Together, these works demonstrate the importance of structured planning, historical context, explicit memory, and task-state management in GUI agent harnesses. While existing studies demonstrate the benefits of particular harness mechanisms, their effects are often studied within different agent systems and evaluation settings. This leaves room for more controlled analysis of how specific harness choices affect performance under a shared environment and how these effects vary with task complexity and the underlying foundation model.

\section{\method Environment}

\subsection{Applications}

\begin{figure}[t]
\centering
\includegraphics[width=\textwidth]{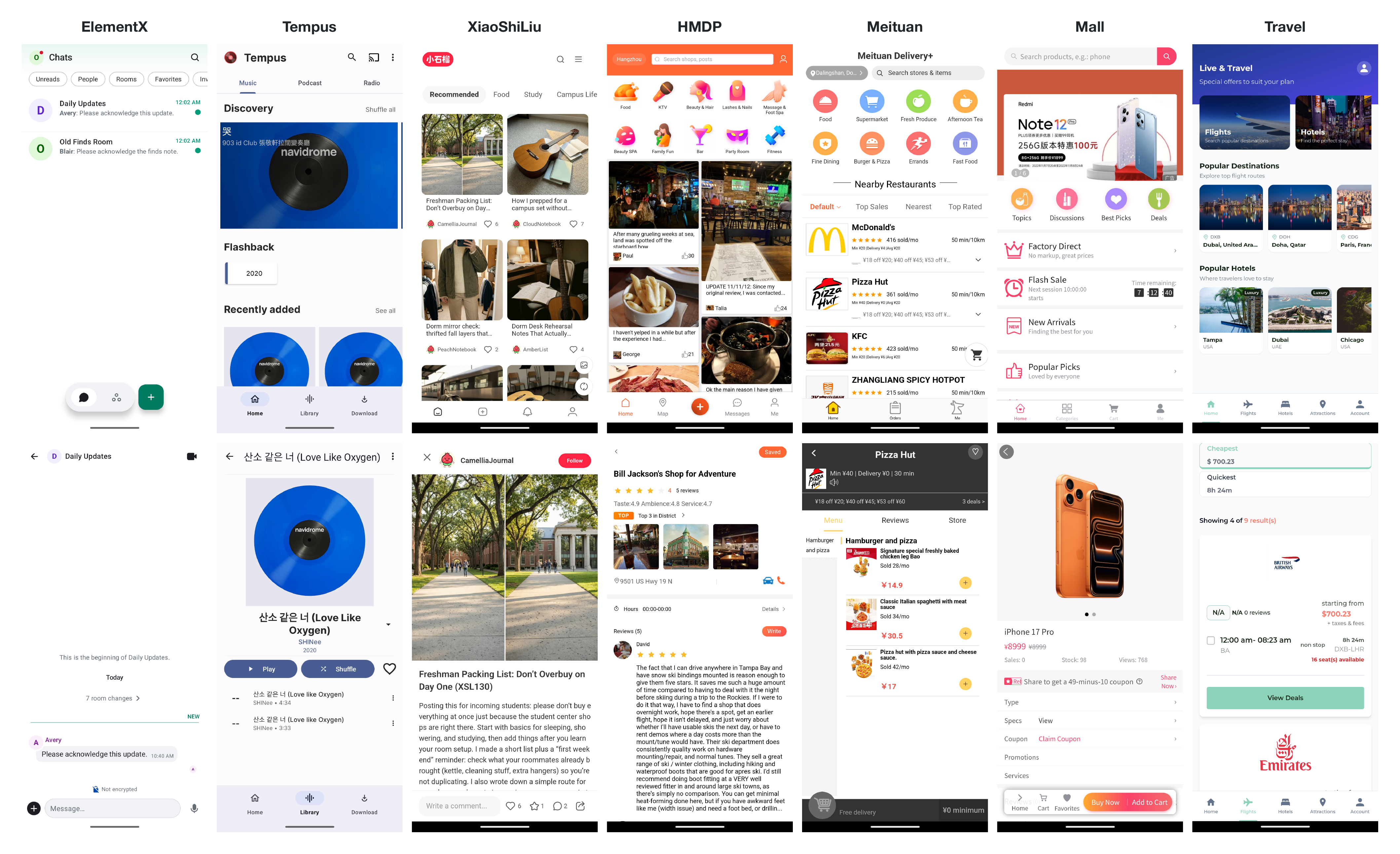}
\vspace{-1em}
\caption{Screenshots of the seven custom applications developed for \method. From left to right, the applications cover instant messaging, music library, lifestyle sharing, local discovery, food delivery, e-commerce, and travel planning, providing diverse interfaces, interaction patterns, and application-specific workflows for mobile agent evaluation.}
\vspace{-1em}
\label{fig:apps}
\end{figure}

To evaluate mobile agents across a broad range of interactions, the underlying environment should cover diverse functional domains, interface structures, and application-specific workflows. Consequently, we expand the benchmark scope by introducing seven custom applications, where each represents a common category of mobile service, deliberately chosen to introduce unique interaction paradigms and challenge distinct facets of the agent reasoning and execution capabilities:

\begin{itemize}[leftmargin=*]
\item \textbf{ElementX\footnote{Android client source: \url{https://github.com/element-hq/element-x-android}. Server source: \url{https://github.com/element-hq/dendrite}.}} serves as a comprehensive instant messaging client built upon the Matrix ecosystem. It supports direct and group conversations, public and private rooms, message editing, file sharing, and interactive polls. This feature density facilitates tasks requiring interpersonal communication, information aggregation, and collaborative coordination.
\item \textbf{Tempus\footnote{Android client source: \url{https://github.com/eddyizm/tempus}. Music server source: \url{https://github.com/navidrome/navidrome}.}} is a music library application linked to a self-hosted Navidrome server. Agents can navigate extensive catalogs of songs, albums, and artists, as well as curate playlists and manage favorites. These mechanics provide an ideal testbed for tasks centered on media retrieval, structured comparison, and the logical organization of digital collections.
\item \textbf{XiaoShiLiu\footnote{Project source: \url{https://github.com/ZTMYO/XiaoShiLiu}.}} models a multimodal lifestyle sharing platform driven by user-generated content. It features categorized images, nested comment threads, likes, content collections, and social following systems. This environment requires agents to parse unstructured social content, extract personalized recommendations, and execute multi-stage social interactions.
\item \textbf{HMDP\footnote{Project source: \url{https://github.com/java-up-up/hmdp-plus}.}} acts as a crowdsourced local discovery and review hub. Agents can search for businesses, scrutinize user reviews, and publish their own experience posts. Additional features like shop favoriting, user following, and voucher management challenge agents to synthesize subjective opinions and navigate localized service ecosystems.
\item \textbf{Meituan\footnote{Web client source: \url{https://github.com/zwStar/vue-meituan}. Server source: \url{https://github.com/zwStar/meituan-backend}.}} replicates an on-demand food delivery service. It encompasses restaurant and menu browsing, address management, cart configurations, ordering workflows, and post-order feedback. Navigating this application requires the agent to handle transactional sequences, spatial constraints, and multi-step checkout procedures.
\item \textbf{Mall\footnote{Web client source: \url{https://github.com/macrozheng/mall-app-web}. Server source: \url{https://github.com/macrozheng/mall}.}} provides a general e-commerce environment. The application supports granular product searches, category browsing, variant selection, shopping cart management, checkout processes, and order tracking. These features rigorously test the capability to navigate hierarchical catalogs, manage user constraints, and confirm commercial transactions.
\item \textbf{Travel\footnote{Project source: \url{https://github.com/mojahidhasan/fullstack-nextjs-golobe-travel-agency}.}} delivers a multi-domain travel planning service that encompasses independent domains for flights, hotels, and tourist attractions. It supports complex search and comparison logic, passenger data entry, secure booking, and payment flows. Navigating these distinct booking pipelines challenges agents with constraint-heavy planning, scheduling dependencies, dense information retrieval, and the meticulous completion of extensive, multi-field forms.
\end{itemize}

While these applications provide the necessary functional breadth, integrating them into a rigorous benchmark requires strict determinism. Rather than relying on simplified mock interfaces, we adapt these applications directly from fully functional open-source projects to preserve authentic interaction flows. To eliminate the variability and unreliability of live external services, we completely decouple all applications from production APIs and we self-host all necessary backend services and databases, populating them with rich, locally managed datasets. Furthermore, we engineer programmatic interfaces to precisely initialize, track, and verify application states during testing, ensuring a fully reproducible evaluation setting.

Beyond these seven custom additions, \method incorporates standard Android system utilities to provide essential foundational functionality. We also retain Mattermost and Mastodon from MobileWorld~\citep{kong2025mobileworld}, supplying enterprise collaboration and decentralized microblogging features. Together, these applications constitute a cohesive and highly diverse mobile ecosystem, and this expansive breadth enables us to design sophisticated, cross-application workflows. Rather than evaluating agents only within isolated applications, \method includes workflows requiring information transfer, state tracking, and sequential execution across functionally distinct domains, thereby capturing important compositional characteristics of practical mobile use.

\subsection{Infrastructure}

\paragraph{Inherited Foundation.}
We build the foundation of \method upon the rigorous evaluation principles and containerized architectures already established by AndroidWorld~\citep{rawles2024androidworlddynamicbenchmarkingenvironment} and MobileWorld~\citep{kong2025mobileworld}. From these frameworks, we inherit fully containerized emulator infrastructures that utilize system snapshots to guarantee reproducible starting environments, alongside objective, state-based verification mechanisms that assess task success without relying on brittle visual heuristics. Extending these robust baselines, our infrastructure development focuses on scaling this environment to accommodate a significantly broader suite of self-hosted applications and the more demanding tasks.

\paragraph{Unified State Abstraction.}
To seamlessly coordinate interactions across these varied functional domains, we introduce a unified, declarative asset abstraction framework. This approach allows heterogeneous application data to become fully composable. An asset represents a typed, modular piece of task state, such as a user contact, message thread, calendar event, social post, e-commerce order, or travel booking. A task specifies its initial conditions as a collection of these assets, which the environment dynamically materializes post-reset via application-specific adapters. This identical representation is then reused to define expected final states, providing a standardized interface for constructing and evaluating tasks across both native on-device applications and our self-hosted services. By decoupling high-level task semantics from low-level database schemas, this design drastically reduces the engineering overhead required to construct our sophisticated scenarios.

\paragraph{Application State Adapters.}
To operationalize this asset framework within our custom applications, we develop dedicated state adapters for each of the seven new functional domains. Every adapter exposes standardized programmatic operations for creating, querying, and resetting evaluation-relevant entities. Crucially, because our self-hosted backends and the Android client interfaces maintain state independently, simply restoring a frontend emulator snapshot is insufficient for full reproducibility. To resolve this discrepancy, our task initialization pipeline synchronizes snapshot restoration with targeted backend resets, cache clearance, session invalidation, and deterministic asset injection.

\paragraph{Composable State Verification.}
Building upon this precise state-alignment infrastructure, we implement a highly reusable verification framework that programmatically tests whether specific assets exist, have been modified, or have been successfully deleted. By defining these state changes as individual criteria, a single task can flexibly combine and weight them, yielding both a strict binary success signal and a fine-grained continuous score for partially completed workflows. Ultimately, this architecture ensures that the exact same objective evaluation interface scales from atomic, single-application actions to long-horizon workflows spanning multiple domains.

\paragraph{Deterministic Text Verification.}
For tasks requiring text entry or explicit answer submission, we intentionally avoid non-deterministic evaluation mechanisms such as semantic matching or LLM-as-a-Judge. Instead, we explicitly specify the expected formatting constraints directly within the task prompt and enforce strict, deterministic verification logic. This design provides stable and reproducible evaluation across models and test runs while avoiding variability introduced by model-based judges. It also imposes a higher operational bar on the agent, testing its ability to adhere strictly to specified formatting contracts and output precise deliverables rather than relying on loose, approximate semantic equivalence.

\subsection{Task Design}

To systematically expose the capability boundaries of current models, we stratify our benchmark into four progressive difficulty tiers based on operation count, cross-application dependencies, information synthesis requirements, and context ambiguity. 

\begin{itemize}[leftmargin=*]
\item \textbf{Atomic tasks} isolate single-step operations with explicitly defined target states, establishing a baseline for GUI element perception and fundamental control execution. 

\item \textbf{Compositional tasks} require chaining multiple operations typically within a single application, evaluating short-horizon planning, basic information retrieval, and state tracking. 

\item \textbf{Cross-application tasks} introduce strict inter-application dependencies, where information synthesized in one environment dictates actions in another, demanding long-horizon planning, working memory, and precise cross-domain information transfer.

\item \textbf{Realistic tasks} represent the most challenging tier, inheriting cross-application dependencies while further introducing underspecified requirements and fragmented evidence across four to seven applications. Agents must parse fragmented evidence amidst distractors, reconcile competing constraints, and execute causally dependent actions. For instance, a typical realistic task requires deducing a product category from an ElementX poll, extracting budget limits from Mail, comparing eligible items in Mall, and scheduling delivery in Calendar.

\end{itemize}

To evaluate this compositionality with rigorous granularity, we structurally decompose each task into one or more discrete subtasks. A subtask represents a specific, independently verifiable milestone within the broader user requirement, such as successfully drafting an email, finalizing a cart checkout, or extracting a precise piece of data. By breaking down complex instructions into a verifiable checklist of these constituent subtasks, the benchmark can explicitly measure partial progress through long-horizon scenarios, directly supporting the continuous scoring metrics defined in our evaluation protocol rather than relying solely on monolithic binary outcomes.

While the difficulty tiers and subtask formulations capture the structural complexity of a workflow, we additionally consider five non-exclusive dimensions during task design. These dimensions provide lightweight guidance for constructing tasks with diverse interaction and reasoning requirements, and also help characterize some of the challenges involved. They are not used as mandatory per-task tags. A task may involve multiple dimensions, while many tasks are not explicitly associated with any of them:

\begin{itemize}[leftmargin=*]
\item \textbf{Multi-Step Workflow} captures tasks that require completing a single overarching objective through a coordinated sequence of actions across multiple apps, where information or outcomes from earlier steps are needed in subsequent steps.

\item \textbf{Selection \& Optimization} captures scenarios where agents must compare multiple feasible alternatives and select the option that best satisfies a set of constraints or preferences.

\item \textbf{Information Gathering} captures tasks that require actively locating, interpreting, and synthesizing information from the environment rather than receiving all necessary parameters directly from the original user instruction. 

\item \textbf{Conditional} captures workflows whose correct execution depends on observed conditions, requiring the agent to select, modify, or terminate a workflow according to the applicable branch.

\item \textbf{Invalid Instruction} captures requests that cannot be completed as stated because they are infeasible, unsupported, or missing essential information, which requires the agent to recognize the limitation and respond appropriately by requesting clarification or proposing a valid alternative.
\end{itemize}

Guided by these criteria, the finalized benchmark comprises a total of 300 meticulously curated tasks, encompassing 73 Atomic, 129 Compositional, 78 Cross-application, and 20 Realistic tasks. To guarantee rigor and reproducibility, every task undergoes a strict dual-annotator quality control process. This review systematically verifies the semantic clarity of the instruction, the accuracy of its assigned difficulty and dimension metadata, and the logical alignment between the intended user requirement and the programmatic verification criteria for each underlying subtask.

\section{Evaluation on \method}

\subsection{Experimental Settings}

\textbf{Action Space.} We anchor our action space in the established MobileWorld framework~\citep{kong2025mobileworld}, adopting its standard GUI operations and navigation commands. To accommodate the demanding evaluation dimensions of \method, we also include \texttt{call\_user} and \texttt{answer}. The \texttt{call\_user} action enables the agent to query a simulated user when critical instruction parameters are missing or ambiguous. Conversely, the \texttt{answer} action allows the agent to explicitly submit text-based deliverables when required by the task instruction. We also include a standard \texttt{terminate} action for the agent to declare that it has finished its execution.

\textbf{Agent Architecture and Context Management.} For our baseline evaluations, we adopt a straightforward, zero-shot autoregressive agent design. Notably, we enforce a strict vision-only observation paradigm where the agent receives only raw screenshots representing the visual state of the device, with zero access to the underlying XML accessibility trees or structural UI hierarchies. Throughout a task, the agent maintains a continuous textual interaction history, accumulating its complete trajectory of reasoning traces and executed actions. To mitigate visual context bloat, we implement a sliding window for visual observations, retaining only the two most recent screenshots while purging older images from the prompt. When the agent utilizes the \texttt{call\_user} action, the response is injected into the subsequent observation state as authoritative context and is permanently retained within the textual history.

\textbf{User Simulation.} To evaluate interactive workflows without human-in-the-loop bottlenecks, we deploy Qwen3.7-Plus~\citep{qwen37plus} as an automated user simulator. To prevent the simulator from inadvertently trivializing the task, it operates under a strict task-specific interaction contract. This prompt-driven contract restricts the simulator to only the information explicitly assigned to the user persona for that scenario, strictly prohibiting it from providing unsolicited operational guidance or revealing step-by-step solutions.

\textbf{Evaluation Protocol and Metrics.} We impose a strict computational budget of 150 interaction steps per episode. An evaluation run terminates under one of three conditions: the agent emits an explicit \texttt{terminate} or \texttt{answer} action, the task verification engine confirms early success, or the maximum step limit is reached. Upon termination, we evaluate the model performance using two primary metrics: \textit{Success} rate and \textit{Average} score. Because tasks frequently comprise multiple constituent subtasks, the strict \textit{Success} rate is a binary metric awarded only when all subtasks within a given task are perfectly satisfied. To capture granular progress on long-horizon workflows, the \textit{Average} score measures the proportion of correctly completed subtasks within a task.

\begin{table*}[t]
    \caption{Overall and difficulty-specific performance of evaluated models on \method. For each setting, we report the strict task success rate (\textit{Success}) and the average proportion of completed subtasks (\textit{Average}). The best result is highlighted in bold, and the second-best result is underlined.}
    \vspace{1em}
    \centering
    \fontsize{7.8pt}{10pt}\selectfont
    \renewcommand{\arraystretch}{1.3}
    \setlength{\tabcolsep}{3.5pt}
    \begin{tabular}{
        @{\hspace{0.03cm}}l|
        cc|cc|cc|cc|cc
        @{\hspace{0.03cm}}
    }
        \toprule
        \multirow{2}{*}{\raisebox{-0.8ex}{\textbf{Models}}}
        & \multicolumn{2}{c|}{\textbf{Overall}}
        & \multicolumn{2}{c|}{\textbf{Atomic}}
        & \multicolumn{2}{c|}{\textbf{Compositional}}
        & \multicolumn{2}{c|}{\textbf{Cross-Application}}
        & \multicolumn{2}{c}{\textbf{Realistic}} \\
        \cmidrule(lr){2-3}
        \cmidrule(lr){4-5}
        \cmidrule(lr){6-7}
        \cmidrule(lr){8-9}
        \cmidrule(lr){10-11}
        & \textit{Success} & \textit{Average}
        & \textit{Success} & \textit{Average}
        & \textit{Success} & \textit{Average}
        & \textit{Success} & \textit{Average}
        & \textit{Success} & \textit{Average} \\
        \midrule

        \textbf{Doubao-Seed-2.1-pro}
        & 62.00 & 74.42
        & \underline{89.04} & \underline{89.73}
        & 65.89 & 79.27
        & 41.03 & 62.99
        & \textbf{20.00} & 31.75 \\
        \midrule

        \textbf{Qwen3.7-Plus}
        & 49.33 & 63.78
        & 78.08 & 80.37
        & 51.94 & 69.78
        & 30.77 & 51.87
        & \phantom{0}0.00 & 11.00 \\

        \textbf{Qwen3.8-Max}
        & \textbf{67.33} & \textbf{80.35}
        & \textbf{93.15} & \textbf{93.84}
        & \textbf{70.54} & \textbf{85.17}
        & \textbf{52.56} & \textbf{70.57}
        & 10.00 & 38.22 \\
        \midrule

        \textbf{Gemini 3.5 Flash}
        & 54.33 & 65.91
        & 80.82 & 81.74
        & 62.02 & 75.23
        & 30.77 & 50.64
        & \phantom{0}0.00 & \phantom{0}7.56 \\
        \midrule

        \textbf{GPT-5.5}
        & 59.67 & 71.14
        & 87.67 & 87.67
        & 65.12 & 78.71
        & 39.74 & 58.78
        & \phantom{0}0.00 & 10.25 \\

        \textbf{GPT-5.6 Sol}
        & \underline{62.67} & \underline{75.86}
        & 86.30 & 86.99
        & \underline{68.99} & \underline{81.60}
        & \underline{43.59} & \underline{64.21}
        & 10.00 & \underline{43.64} \\
        \midrule

        \textbf{Claude Sonnet 4.6}
        & 55.33 & 68.13
        & 83.56 & 83.56
        & 60.47 & 75.27
        & 34.62 & 50.33
        & \phantom{0}0.00 & 35.06 \\

        \textbf{Claude Opus 4.7}
        & 60.00 & 72.33
        & 87.67 & 87.67
        & 65.12 & 78.01
        & 37.18 & 55.54
        & \underline{15.00} & \textbf{45.27} \\

        \bottomrule
    \end{tabular}
    \label{tab:main_result}
\end{table*}

\subsection{Results and Analyses}

We evaluate eight frontier models from five model families: Doubao-Seed-2.1-pro~\citep{bytedance2026seed21} from ByteDance Seed; Qwen3.7-Plus~\citep{qwen37plus} and Qwen3.8-Max~\citep{qwen38} from Alibaba Qwen; Gemini 3.5 Flash~\citep{googledeepmind2026gemini35flash} from Google DeepMind; GPT-5.5~\citep{openai2026gpt55} and GPT-5.6 Sol~\citep{openai2026gpt56} from OpenAI; Claude Sonnet 4.6~\citep{anthropic2026claudesonnet46} and Claude Opus 4.7~\citep{anthropic2026claudeopus47} from Anthropic. Table~\ref{tab:main_result} reports both \textit{Success} rate and \textit{Average} score for each model across the four difficulty tiers.

\textbf{Overall performance.}
Qwen3.8-Max achieves the strongest overall performance, reaching a \textit{Success} rate of 67.33 and an \textit{Average} score of 80.35. It also ranks first on both metrics across the atomic, compositional, and cross-application tiers, demonstrating consistently strong performance on tasks up to substantial compositional complexity. GPT-5.6 Sol emerges as the strongest runner-up, achieving the second-best overall \textit{Success} rate of 62.67 and \textit{Average} score of 75.86. It also maintains competitive performance across different difficulty tiers, consistently ranking among the top-performing models as task complexity increases. Doubao-Seed-2.1-pro follows closely, obtaining an overall \textit{Success} rate of 62.00 and an \textit{Average} score of 74.42.

\textbf{Performance across difficulty levels.}
All models exhibit a clear performance decline as task difficulty increases. On atomic tasks, the strongest model achieves 93.15 and 93.84 in \textit{Success} and \textit{Average} respectively, and all models obtain \textit{Average} scores above 80. Performance decreases considerably on cross-application tasks, where Qwen3.8-Max reaches only 52.56 in \textit{Success}. This consistent degradation indicates that composing more operations, preserving intermediate information, and managing complex dependencies remain challenging even for the strongest models.

\textbf{Challenges in realistic workflows.}
The decline becomes particularly pronounced on the realistic tier. No model completes more than 20\% of these tasks perfectly, with Doubao-Seed-2.1-pro achieving the highest \textit{Success} rate of 20.00. Claude Opus 4.7 achieves the highest \textit{Average} score of 45.27, followed by GPT-5.6 Sol with 43.64, while the overall leader Qwen3.8-Max obtains 38.22. The substantial gap shows that models can often complete individual parts of these workflows but struggle to satisfy all interdependent requirements within a single episode.

\textbf{Different strengths on complex tasks.}
The model ranking on realistic tasks differs notably from the overall performance. Doubao-Seed-2.1-pro achieves the highest \textit{Success} rate, while Claude Opus 4.7 achieves the highest \textit{Average} score, despite neither model leading the overall results. Conversely, Qwen3.8-Max dominates the first three difficulty tiers but does not retain the same advantage on realistic tasks. These results suggest that strong performance on atomic and moderately compositional interactions does not necessarily translate to reliable execution of holistic mobile workflows involving long-horizon reasoning, information synthesis, and multiple cross-application dependencies.

\subsection{Performance Across Applications}

\begin{wrapfigure}{r}{0.50\textwidth}
    \centering
    \vspace{-1.2em}
    \includegraphics[width=0.5\textwidth]{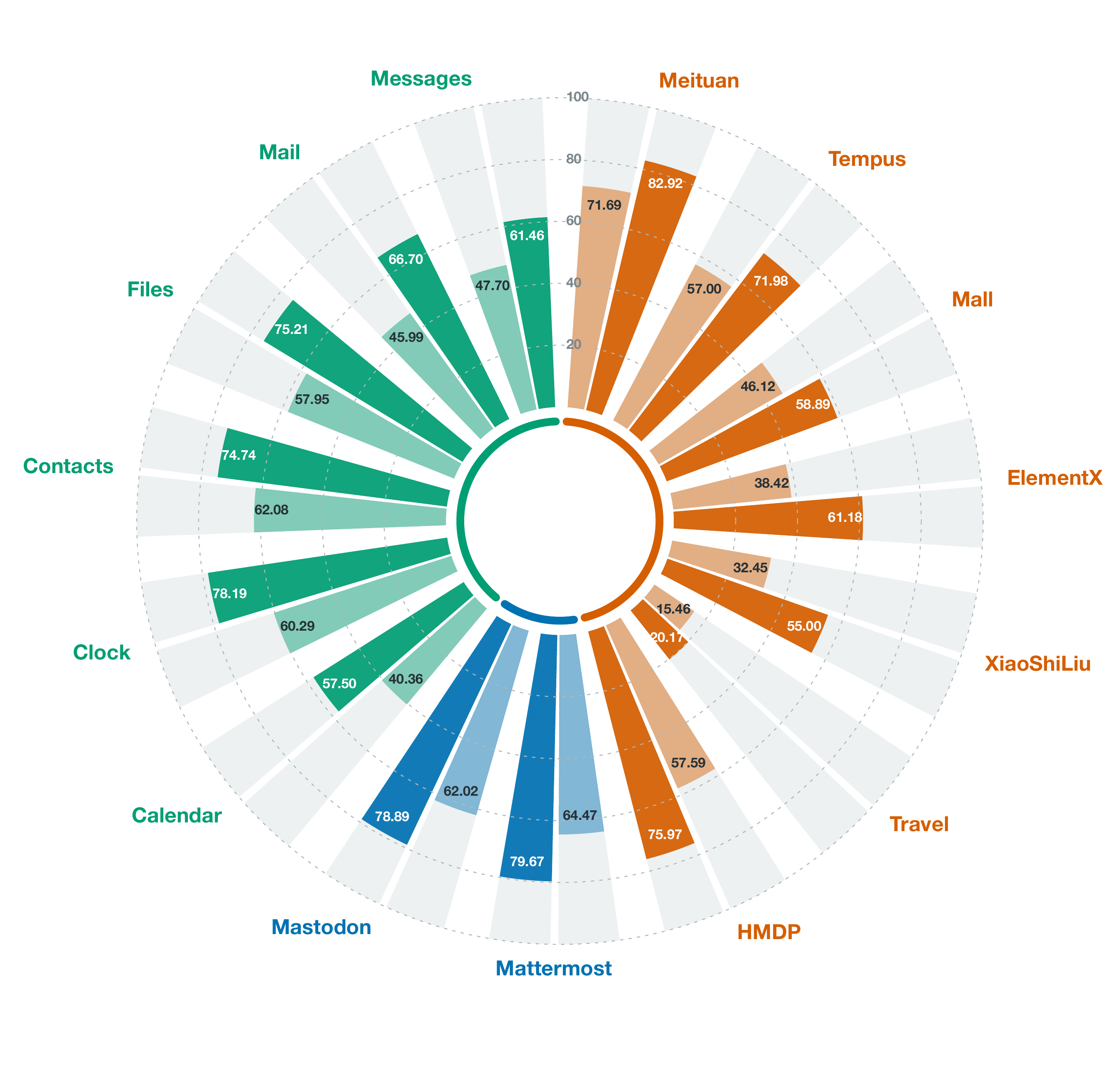}
    \caption{Performance across different applications, averaged over all evaluated models.
    Light and dark bars denote \textit{Success} rate and \textit{Average} scores, respectively.
    Orange denotes applications introduced in \method, blue denotes applications inherited from MobileWorld, and green denotes Android system applications.}
    \label{fig:app_performance}
    \vspace{-1.2em}
\end{wrapfigure}

We further examine how agent performance varies across the application ecosystem, and Figure~\ref{fig:app_performance} reveals substantial variation. Travel is particularly challenging, with an \textit{Average} score of only 20.17, followed by XiaoShiLiu at 55.00. Several other applications introduced in \method, including Mall and ElementX, also fall toward the lower end of the performance range. Meanwhile, some newly introduced applications, such as Meituan and HMDP, achieve substantially stronger results. These observations suggest that the difficulty introduced by broader application coverage is not uniform, but several of the weakest-performing scenarios do arise from applications newly covered by our benchmark.

This variation highlights an important challenge in evaluating mobile agents across a broader application ecosystem. Strong performance on commonly evaluated applications does not necessarily translate uniformly to all additional scenarios, and current agents can still experience substantial degradation when interacting with different application interfaces, workflows, and task settings. The low performance observed on several newly introduced applications therefore provides evidence that current capabilities have not yet generalized reliably across the full range of mobile environments considered in \method. Expanding application coverage is thus important not only for more comprehensive evaluation, but also for developing and training mobile agents that can generalize more reliably across diverse interfaces and usage scenarios.

\section{Harness Matters for \method Tasks}

Harness design has emerged as an important topic in agent research, with growing evidence that an effective harness can substantially unlock model capabilities in domains such as software engineering. However, comparatively little work has examined how harness design could influence performance for mobile agents, especially when tasks require sustained interaction over long and complex trajectories. This motivates us to take a closer look at the role of harness design in \method, with the goal of understanding not only whether different mechanisms help, but also how their effects vary and interact under increasingly demanding mobile tasks.

\subsection{Context Retention}

We first study how the amount of retained visual context affects agent performance. As shown in Table~\ref{tab:context_retention}, increasing the number of historical screenshots generally improves performance, especially when moving from a very limited visual context to a moderate one. With only one screenshot retained, Qwen3.7-Plus achieves an overall \textit{Success} rate of 43.33 and \textit{Average} score of 57.20, while retaining eight screenshots increases these scores to 52.00 and 67.10 respectively. The improvement is also pronounced on realistic tasks, where the average score rises from 9.97 to 21.44. However, the marginal benefit gradually diminishes as more screenshots are added, and retaining 16 images does not bring further overall improvement, with the \textit{Average} score slightly decreasing. This suggests that historical visual observations are useful for preserving task-relevant information, but simply enlarging the visual context is insufficient once a moderate amount of history is already available.

\begin{wraptable}{r}{0.50\textwidth}
    \vspace{-2.0em}
    \caption{Effect of context retention on Qwen3.7-Plus. \textbf{Images} denotes the number of most recent screenshots retained in the context, \textbf{Thought} indicates whether previous reasoning traces are preserved, and \textbf{Caption} indicates whether older screenshots are retained as captions.}
    \label{tab:context_retention}
    \vspace{0.5em}
    \centering

    \fontsize{7.5pt}{9pt}\selectfont
    \renewcommand{\arraystretch}{1.2}
    \setlength{\tabcolsep}{2.5pt}

    \resizebox{\linewidth}{!}{
    \begin{tabular}{
        @{\hspace{0.02cm}}ccc|
        cc|cccc
        @{\hspace{0.02cm}}
    }
        \toprule
        \multirow{2}{*}{\raisebox{-0.8ex}{\textbf{Images}}}
        & \multirow{2}{*}{\raisebox{-0.8ex}{\textbf{Thought}}}
        & \multirow{2}{*}{\raisebox{-0.8ex}{\textbf{Caption}}}
        & \multicolumn{2}{c|}{\textbf{Overall}}
        & \multicolumn{4}{c}{\textbf{Difficulty}} \\
        \cmidrule(lr){4-5}
        \cmidrule(lr){6-9}
        & & & \textit{Success} & \textit{Average}
        & \textit{Atomic}
        & \textit{Comp.}
        & \textit{Cross-App.}
        & \textit{Realistic} \\
        \midrule

        1  & \cmark & \xmark
        & 43.33 & 57.20
        & 72.83 & 61.67 & 47.30 & \phantom{0}9.97 \\

        2  & \cmark & \xmark
        & 49.33 & 63.78
        & 80.37 & 69.78 & 51.87 & 11.00 \\

        2  & \xmark & \xmark
        & 41.67 & 55.77
        & 80.37 & 57.13 & 43.03 & \phantom{0}6.89 \\

        2  & \cmark & \cmark
        & 51.33 & 65.13
        & 78.31 & 70.65 & 51.50 & 34.60 \\

        4  & \cmark & \xmark
        & 51.00 & 65.53
        & 80.37 & 71.89 & 54.64 & 12.86 \\

        8  & \cmark & \xmark
        & 52.00 & 67.10
        & 83.56 & 72.32 & 54.77 & 21.44 \\

        16 & \cmark & \xmark
        & 52.00 & 65.52
        & 79.68 & 70.95 & 53.97 & 23.81 \\

        \bottomrule
    \end{tabular}
    }
\end{wraptable}

Retaining previous reasoning traces provides another important source of historical information. Under the same two-image setting, removing previous thoughts decreases the \textit{Success} rate from 49.33 to 41.67 and the \textit{Average} score from 63.78 to 55.77. With only a small number of screenshots retained, removing the reasoning history effectively discards most information about earlier interaction steps, leaving the agent to rely largely on the current interface state and a very limited visual history. The impact becomes more evident on more complex tasks. \textit{Average} performance drops from 69.78 to 57.13 on compositional tasks, from 51.87 to 43.03 on cross-application tasks, and from 11.00 to 6.89 on realistic tasks. We further examine an alternative strategy that replaces discarded historical screenshots with textual captions. Under the two-image setting, retaining captions for older screenshots improves the overall \textit{Success} rate from 49.33 to 51.33 and the \textit{Average} score from 63.78 to 65.13. The improvement is particularly substantial on realistic tasks, where the \textit{Average} score increases from 11.00 to 34.60. This suggests that compact textual summaries of older visual observations can preserve useful historical information without requiring all screenshots to remain in the context. Together, these results suggest that maintaining appropriate historical context is particularly important for tasks that require agents to preserve intermediate information and track progress over long interaction trajectories.

\subsection{Explicit State Tracking}

Modern agents commonly maintain explicit memory or task status to preserve critical information across long interaction trajectories. For mobile agents, this becomes particularly important as relevant information may disappear from the current screen while task progress accumulates across multiple actions and applications. To study the effect of such structured memory, we implement an explicit state tracking mechanism with reference to the Task-State Representation (TSR) framework proposed in ~\cite{zheng2026taskstaterepresentationlonghorizonmobile}, maintaining task requirements, subgoal progress, and execution state throughout the trajectory. We compare agents with and without this mechanism under different amounts of retained visual history.

\begin{wrapfigure}{r}{0.50\textwidth}
\centering
\includegraphics[width=0.6\textwidth]{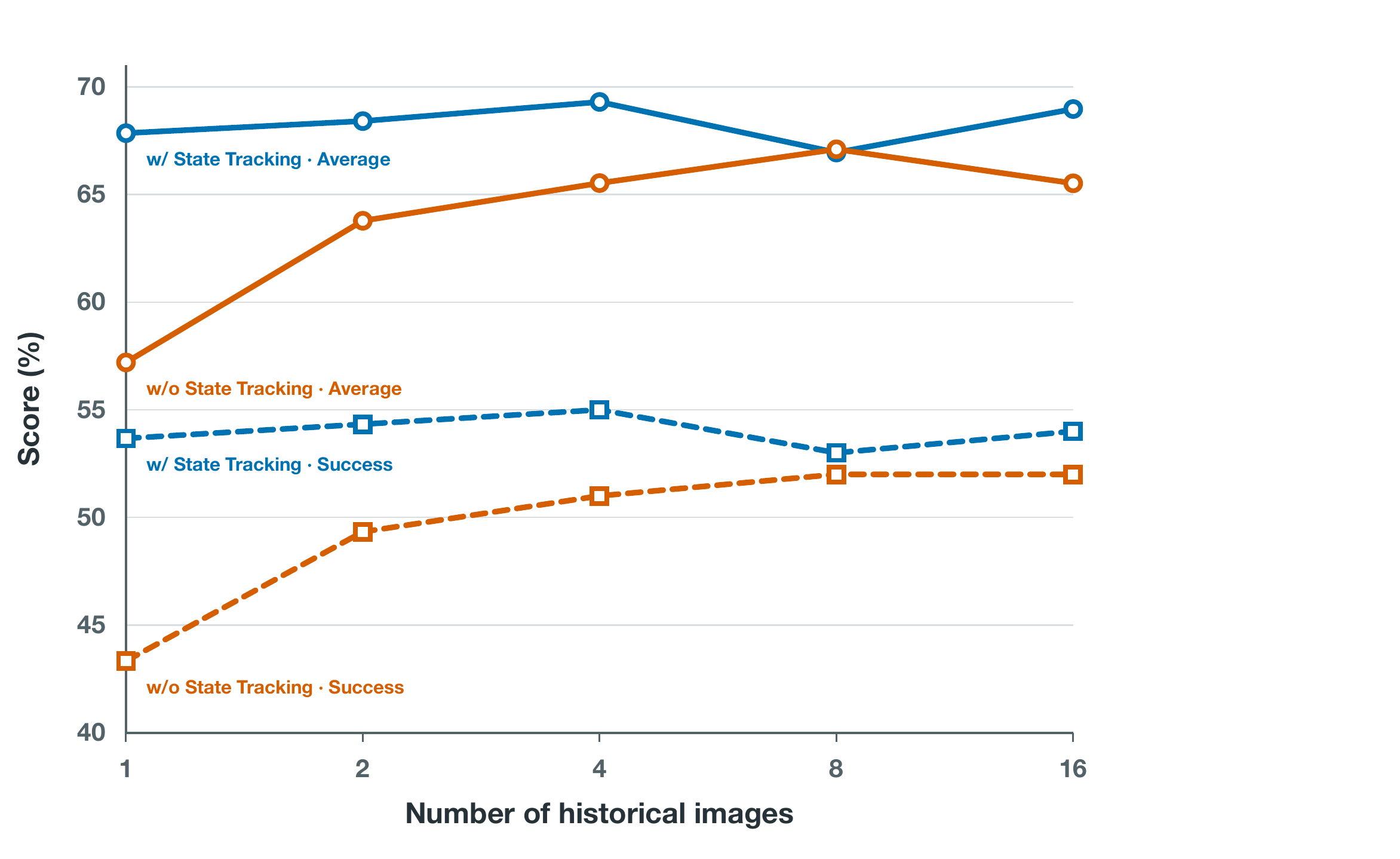}
\caption{Overall performance with and without explicit state tracking under different numbers of retained screenshots.}
\label{fig:state_tracking_overall}
\vspace{-1.0em}
\end{wrapfigure}

Figure~\ref{fig:state_tracking_overall} shows that explicit state tracking generally improves agent performance across different visual context sizes. The improvement is particularly clear when visual history is limited. With one retained screenshot, state tracking increases \textit{Success} rate from 43.33 to 53.67 and the \textit{Average} score from 57.20 to 67.85. More broadly, it improves the \textit{Success} rate under all tested context sizes and improves the \textit{Average} score in most settings. The benefit becomes smaller as more screenshots are retained, and with eight screenshots the \textit{Average} score remains nearly unchanged, decreasing slightly from 67.10 to 66.95. These results indicate that explicitly maintaining task state can provide useful information beyond simply retaining raw interaction history, particularly when the available visual history is limited.

\begin{figure}[t]
\centering
\includegraphics[width=\textwidth]{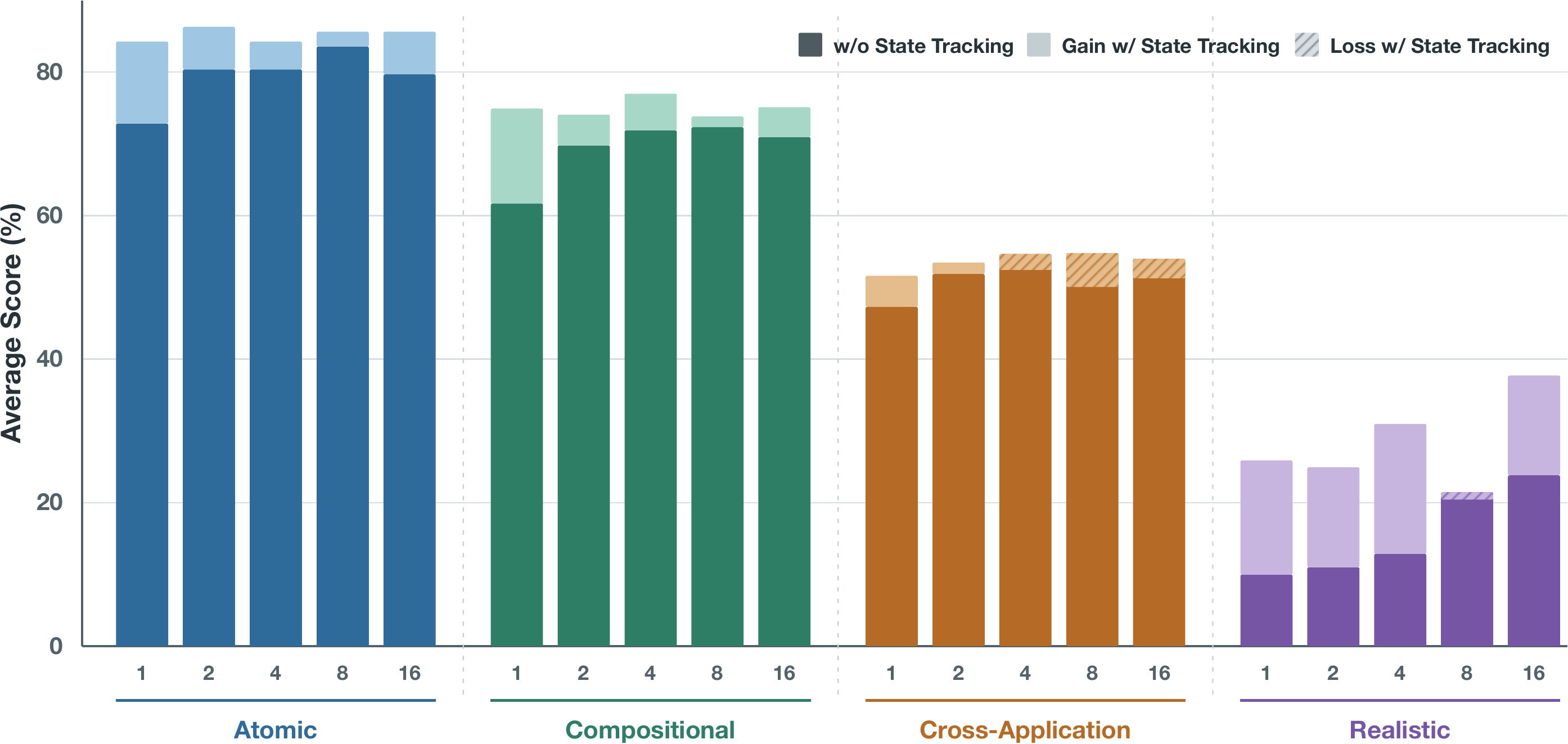}
\caption{Effect of state tracking across different difficulty tiers. The solid portion denotes the \textit{Average} score without state tracking, while the lighter or hatched portion indicates the gain or loss after enabling state tracking. Numbers below each group denote the number of retained screenshots.}
\label{fig:state_tracking_difficulty}
\end{figure}

Results in Figure~\ref{fig:state_tracking_difficulty} show that the effect of state tracking differs substantially across difficulty tiers. On atomic and compositional tasks, state tracking consistently improves the \textit{Average} score across all tested visual context sizes. For example, with one retained screenshot, the score increases from 72.83 to 84.25 on atomic tasks and from 61.67 to 74.90 on compositional tasks. Positive gains remain even when more screenshots are retained, indicating that explicit task state provides useful information beyond raw visual history for these tasks. The effect is less consistent on cross-application tasks. State tracking improves performance when only one or two screenshots are retained, but slightly decreases the \textit{Average} score with four, eight, and sixteen screenshots. In contrast, the strongest benefits appear on realistic tasks. State tracking increases the \textit{Average} score from 9.97 to 25.89 with one screenshot, from 12.86 to 30.97 with four screenshots, and from 23.81 to 37.72 even with sixteen screenshots. The eight-image setting is the only exception, where performance decreases slightly from 21.44 to 20.42. Overall, these results suggest that explicitly organizing task requirements and execution progress is particularly valuable for realistic workflows.

The amount of retained visual context further modulates the benefit of state tracking. For the first three difficulty tiers, the improvement brought by state tracking generally becomes smaller as more historical screenshots are retained, suggesting that a richer visual context can already preserve part of the information that explicit state tracking provides. This trend is particularly evident compared with the large gains observed under the one-image setting. Realistic tasks, however, exhibit a notably different pattern. Even when sixteen screenshots are retained, state tracking still increases the \textit{Average} score from 23.81 to 37.72, corresponding to a substantial gain of 13.91 points, which is only slightly smaller than the 15.92-point improvement observed with a single retained screenshot. This suggests that, for highly complex workflows, explicit state tracking can still provide valuable complementary information even when rich visual history is already available, by organizing task requirements, intermediate progress, and cross-step dependencies in a more accessible form.

\subsection{Model-Specific Harness Design}

The effectiveness of a harness may also depend on the underlying foundation model. To examine this assumption, we compare two state-tracking designs on Qwen3.7-Plus and GPT-5.6 Sol. The \textit{Free-form} design allows the model to maintain task state in an unconstrained textual format, whereas the \textit{Structured} design used in our previous experiments requires the model to update a predefined set of state fields. We additionally report the performance without explicit state tracking as a reference.

The two models respond quite differently to the same state-tracking designs. For Qwen3.7-Plus, the structured representation provides a clear advantage, increasing the \textit{Success} rate from 49.33 to 54.33 and the \textit{Average} score from 63.78 to 68.41. The improvement is particularly large on realistic tasks, where the score rises from 11.00 without state tracking to 24.92. In comparison, the free-form representation brings only limited overall improvement and even performs worse on realistic tasks. This suggests that Qwen3.7-Plus benefits from an explicit structure that guides how task-relevant information and progress are maintained.

\begin{wraptable}{r}{0.50\textwidth}
    \vspace{-2.0em}
    \caption{Effect of different state-tracking designs across models.
    \textit{Free-form} allows unconstrained state representation, while
    \textit{Structured} uses predefined state fields.}
    \label{tab:model_specific_harness}
    \vspace{1em}
    \centering

    \fontsize{7.5pt}{9pt}\selectfont
    \renewcommand{\arraystretch}{1.2}
    \setlength{\tabcolsep}{2.3pt}

    \resizebox{\linewidth}{!}{
    \begin{tabular}{
        @{\hspace{0.02cm}}ll|
        cc|cccc
        @{\hspace{0.02cm}}
    }
        \toprule
        \multirow{2}{*}{\raisebox{-0.8ex}{\textbf{Model}}}
        & \multirow{2}{*}{\raisebox{-0.8ex}{\textbf{State}}}
        & \multicolumn{2}{c|}{\textbf{Overall}}
        & \multicolumn{4}{c}{\textbf{Difficulty}} \\
        \cmidrule(lr){3-4}
        \cmidrule(lr){5-8}
        & & \textit{Success} & \textit{Average}
        & \textit{Atomic}
        & \textit{Comp.}
        & \textit{Cross-App.}
        & \textit{Realistic} \\
        \midrule

        \multirow{3}{*}{\textbf{Qwen}}
        & None
        & 49.33 & 63.78
        & 80.37 & 69.78 & 51.87 & 11.00 \\

        & Free-form
        & 50.33 & 64.73
        & 85.16 & 71.27 & 49.37 & \phantom{0}7.89 \\

        & Structured
        & 54.33 & 68.41
        & 86.30 & 74.07 & 53.46 & 24.92 \\

        \midrule

        \multirow{3}{*}{\textbf{GPT}}
        & None
        & 62.67 & 75.86
        & 86.99 & 81.60 & 64.21 & 43.64 \\

        & Free-form
        & 65.67 & 75.75
        & 87.67 & 84.11 & 60.32 & 38.54 \\

        & Structured
        & 64.00 & 73.96
        & 88.36 & 82.72 & 54.04 & 42.56 \\

        \bottomrule
    \end{tabular}
    }
    \vspace{-1.0em}
\end{wraptable}

The same pattern does not hold for GPT-5.6 Sol. The free-form representation achieves the highest overall \textit{Success} rate of 65.67, but its \textit{Average} score of 75.75 remains nearly unchanged from 75.86 without state tracking. The structured representation achieves a \textit{Success} rate of 64.00 while reducing the overall \textit{Average} score to 73.96. Its effect also varies across difficulty tiers. The structured design slightly improves performance on atomic and compositional tasks, but decreases the \textit{Average} score on cross-application tasks from 64.21 to 54.04 and on realistic tasks from 43.64 to 42.56. Thus, unlike Qwen3.7-Plus, GPT-5.6 Sol does not consistently benefit from adding explicit structure to its state representation.

These results together suggest that harness mechanisms are not universally transferable across foundation models. Additional structure can effectively guide one model while unnecessarily constraining another. Harness design should therefore consider the characteristics of the underlying model rather than assuming a single configuration is optimal for all agents. We view this analysis as an initial investigation, and broader experiments across more foundation models, harness designs, and task settings are still needed to better characterize these model-harness interactions and examine how consistently they generalize across mobile scenarios.

\section{Conclusion}

We present \method, a benchmark for evaluating general mobile assistants in challenging mobile scenarios, featuring seven custom applications, 300 tasks across four progressive difficulty tiers, and a reproducible evaluation infrastructure. Evaluation of eight frontier models shows a clear performance decline as task complexity increases, with cross-application and realistic workflows remaining particularly challenging. We further study several agent harness choices and find that context retention and explicit state tracking can meaningfully affect performance, while the same design may not benefit different foundation models equally. Overall, \method provides a broader and more challenging testbed for evaluating mobile agents and studying how model capabilities and harness design shape performance in complex mobile environments.

\bibliography{iclr2025_conference}

@misc{wei2025browsecompsimplechallengingbenchmark,
      title={BrowseComp: A Simple Yet Challenging Benchmark for Browsing Agents}, 
      author={Jason Wei and Zhiqing Sun and Spencer Papay and Scott McKinney and Jeffrey Han and Isa Fulford and Hyung Won Chung and Alex Tachard Passos and William Fedus and Amelia Glaese},
      year={2025},
      eprint={2504.12516},
      archivePrefix={arXiv},
      primaryClass={cs.CL},
      url={https://arxiv.org/abs/2504.12516}, 
}

@misc{merrill2026terminalbenchbenchmarkingagentshard,
      title={Terminal-Bench: Benchmarking Agents on Hard, Realistic Tasks in Command Line Interfaces},
      author={Mike A. Merrill and Alexander G. Shaw and Nicholas Carlini and Boxuan Li and Harsh Raj and Ivan Bercovich and Lin Shi and Jeong Yeon Shin and Thomas Walshe and E. Kelly Buchanan and Junhong Shen and Guanghao Ye and Haowei Lin and Jason Poulos and Maoyu Wang and Marianna Nezhurina and Jenia Jitsev and Di Lu and Orfeas Menis Mastromichalakis and Zhiwei Xu and Zizhao Chen and Yue Liu and Robert Zhang and Leon Liangyu Chen and Anurag Kashyap and Jan-Lucas Uslu and Jeffrey Li and Jianbo Wu and Minghao Yan and Song Bian and Vedang Sharma and Ke Sun and Steven Dillmann and Akshay Anand and Andrew Lanpouthakoun and Bardia Koopah and Changran Hu and Etash Guha and Gabriel H. S. Dreiman and Jiacheng Zhu and Karl Krauth and Li Zhong and Niklas Muennighoff and Robert Amanfu and Shangyin Tan and Shreyas Pimpalgaonkar and Tushar Aggarwal and Xiangning Lin and Xin Lan and Xuandong Zhao and Yiqing Liang and Yuanli Wang and Zilong Wang and Changzhi Zhou and David Heineman and Hange Liu and Harsh Trivedi and John Yang and Junhong Lin and Manish Shetty and Michael Yang and Nabil Omi and Negin Raoof and Shanda Li and Terry Yue Zhuo and Wuwei Lin and Yiwei Dai and Yuxin Wang and Wenhao Chai and Shang Zhou and Dariush Wahdany and Ziyu She and Jiaming Hu and Zhikang Dong and Yuxuan Zhu and Sasha Cui and Ahson Saiyed and Arinbjörn Kolbeinsson and Jesse Hu and Christopher Michael Rytting and Ryan Marten and Yixin Wang and Alex Dimakis and Andy Konwinski and Ludwig Schmidt},
      year={2026},
      eprint={2601.11868},
      archivePrefix={arXiv},
      primaryClass={cs.SE},
      url={https://arxiv.org/abs/2601.11868},
}

@misc{datacurve2026deepswev11,
  title  = {DeepSWE v1.1: a cleaner, more reproducible benchmark for frontier coding agents},
  author = {Wenqi Huang and Peter Jiang},
  year   = {2026},
  url    = {https://github.com/datacurve-ai/deep-swe},
}

@misc{OSWorld,
      title={OSWorld: Benchmarking Multimodal Agents for Open-Ended Tasks in Real Computer Environments}, 
      author={Tianbao Xie and Danyang Zhang and Jixuan Chen and Xiaochuan Li and Siheng Zhao and Ruisheng Cao and Toh Jing Hua and Zhoujun Cheng and Dongchan Shin and Fangyu Lei and Yitao Liu and Yiheng Xu and Shuyan Zhou and Silvio Savarese and Caiming Xiong and Victor Zhong and Tao Yu},
      year={2024},
      eprint={2404.07972},
      archivePrefix={arXiv},
      primaryClass={cs.AI}
}

@misc{yuan2026osworld20benchmarkingcomputeruse,
      title={OSWorld2.0: Benchmarking Computer Use Agents on Long-Horizon Real-World Tasks},
      author={Mengqi Yuan and Zilong Zhou and Xinzhuang Xiong and Weiming Wu and Jiayang Sun and Jiamin Song and Kaiqian Cui and Bowen Wang and Haoyuan Wu and Yitong Li and Dunjie Lu and Haikong Lu and Qi Zhen and Xinyuan Wang and Jiaqi Deng and Yuhao Yang and Cheng Chen and Boyuan Zheng and Alex Su and Xiao Yu and Hao Zou and Saaket Agashe and Xing Han Lu and Manpreet Kaur and Zhengyang Qi and Vincent Sunn Chen and Frederic Sala and Dayiheng Liu and Junyang Lin and Zhou Yu and Yu Su and Siva Reddy and Xin Eric Wang and Peng Qi and Tianbao Xie and Tao Yu},
      year={2026},
      eprint={2606.29537},
      archivePrefix={arXiv},
      primaryClass={cs.AI},
      url={https://arxiv.org/abs/2606.29537},
}

@misc{rawles2024androidworlddynamicbenchmarkingenvironment,
      title={AndroidWorld: A Dynamic Benchmarking Environment for Autonomous Agents},
      author={Christopher Rawles and Sarah Clinckemaillie and Yifan Chang and Jonathan Waltz and Gabrielle Lau and Marybeth Fair and Alice Li and William Bishop and Wei Li and Folawiyo Campbell-Ajala and Daniel Toyama and Robert Berry and Divya Tyamagundlu and Timothy Lillicrap and Oriana Riva},
      year={2024},
      eprint={2405.14573},
      archivePrefix={arXiv},
      primaryClass={cs.AI},
      url={https://arxiv.org/abs/2405.14573},
}

@inproceedings{kong2025mobileworld,
      title={MobileWorld: Benchmarking Autonomous Mobile Agents in Agent-User Interactive, and MCP-Augmented Environments},
      author={Quyu Kong and Xu Zhang and Zhenyu Yang and Nolan Gao and Chen Liu and Panrong Tong and Chenglin Cai and Hanzhang Zhou and Jianan Zhang and Liangyu Chen and Zhidan Liu and Steven Hoi and Yue Wang},
      booktitle={Proceedings of the 64th Annual Meeting of the Association for Computational Linguistics (ACL)},
      year={2026},
      url={https://arxiv.org/abs/2512.19432},
}

@inproceedings{li2025mobileuse,
  title={MobileUse: A Hierarchical Reflection-Driven {GUI} Agent for Autonomous Mobile Operation},
  author={Ning Li and Xiangmou Qu and Jiamu Zhou and Jun Wang and Muning Wen and Kounianhua Du and Xingyu Lou and Qiuying Peng and Jun Wang and Weinan Zhang},
  booktitle={The Thirty-ninth Annual Conference on Neural Information Processing Systems},
  year={2025},
  url={https://openreview.net/forum?id=KR6tnkb6h4}
}

@misc{shi2026androtmeminteractiontrajectoriesanchored,
      title={AndroTMem: From Interaction Trajectories to Anchored Memory in Long-Horizon GUI Agents}, 
      author={Yibo Shi and Jungang Li and Linghao Zhang and Zihao Dongfang and Biao Wu and Sicheng Tao and Yibo Yan and Chenxi Qin and Weiting Liu and Zhixin Lin and Hanqian Li and Yu Huang and Song Dai and Yonghua Hei and Yue Ding and Xiang Li and Shikang Wang and Chengdong Xu and Jingqi Liu and Xueying Ma and Zhiwen Zheng and Xiaofei Zhang and Bincheng Wang and Nichen Yang and Jie Wu and Lihua Tian and Chen Li and Xuming Hu},
      year={2026},
      eprint={2603.18429},
      archivePrefix={arXiv},
      primaryClass={cs.CV},
      url={https://arxiv.org/abs/2603.18429}, 
}

@inproceedings{NEURIPS2022_82ad13ec,
 author = {Yao, Shunyu and Chen, Howard and Yang, John and Narasimhan, Karthik},
 booktitle = {Advances in Neural Information Processing Systems},
 doi = {10.52202/068431-1508},
 editor = {S. Koyejo and S. Mohamed and A. Agarwal and D. Belgrave and K. Cho and A. Oh},
 pages = {20744--20757},
 publisher = {Curran Associates, Inc.},
 title = {WebShop: Towards Scalable Real-World Web Interaction with Grounded Language Agents},
 url = {https://proceedings.neurips.cc/paper_files/paper/2022/file/82ad13ec01f9fe44c01cb91814fd7b8c-Paper-Conference.pdf},
 volume = {35},
 year = {2022}
}

@article{zhou2023webarena,
  title={WebArena: A Realistic Web Environment for Building Autonomous Agents},
  author={Zhou, Shuyan and Xu, Frank F and Zhu, Hao and Zhou, Xuhui and Lo, Robert and Sridhar, Abishek and Cheng, Xianyi and Bisk, Yonatan and Fried, Daniel and Alon, Uri and others},
  journal={arXiv preprint arXiv:2307.13854},
  year={2023}
}

@inproceedings{
    jimenez2024swebench,
    title={{SWE}-bench: Can Language Models Resolve Real-world Github Issues?},
    author={Carlos E Jimenez and John Yang and Alexander Wettig and Shunyu Yao and Kexin Pei and Ofir Press and Karthik R Narasimhan},
    booktitle={The Twelfth International Conference on Learning Representations},
    year={2024},
    url={https://openreview.net/forum?id=VTF8yNQM66}
}

@misc{deng2025swebenchproaiagents,
      title={SWE-Bench Pro: Can AI Agents Solve Long-Horizon Software Engineering Tasks?}, 
      author={Xiang Deng and Jeff Da and Edwin Pan and Yannis Yiming He and Charles Ide and Kanak Garg and Niklas Lauffer and Andrew Park and Nitin Pasari and Chetan Rane and Karmini Sampath and Maya Krishnan and Srivatsa Kundurthy and Sean Hendryx and Zifan Wang and Vijay Bharadwaj and Jeff Holm and Raja Aluri and Chen Bo Calvin Zhang and Noah Jacobson and Bing Liu and Brad Kenstler},
      year={2025},
      eprint={2509.16941},
      archivePrefix={arXiv},
      primaryClass={cs.SE},
      url={https://arxiv.org/abs/2509.16941}, 
}

@misc{yang2026programbenchlanguagemodelsrebuild,
      title={ProgramBench: Can Language Models Rebuild Programs From Scratch?},
      author={John Yang and Kilian Lieret and Jeffrey Ma and Parth Thakkar and Dmitrii Pedchenko and Sten Sootla and Emily McMilin and Pengcheng Yin and Rui Hou and Gabriel Synnaeve and Diyi Yang and Ofir Press},
      year={2026},
      eprint={2605.03546},
      archivePrefix={arXiv},
      primaryClass={cs.SE},
      url={https://arxiv.org/abs/2605.03546},
}

@article{du2025deepresearch,
  author    = {Mingxuan Du and Benfeng Xu and Chiwei Zhu and Xiaorui Wang and Zhendong Mao},
  title     = {DeepResearch Bench: A Comprehensive Benchmark for Deep Research Agents},
  journal   = {arXiv preprint},
  year      = {2025},
}

@misc{tang2026workspacebench10benchmarkingai,
      title={Workspace-Bench 1.0: Benchmarking AI Agents on Workspace Tasks with Large-Scale File Dependencies}, 
      author={Zirui Tang and Xuanhe Zhou and Yumou Liu and Linchun Li and Weizheng Wang and Hongzhang Huang and Jun Zhou and Jiachen Song and Shaoli Yu and Jinqi Wang and Zihang Zhou and Hongyi Zhou and Yuting Lv and Jinyang Li and Jiashuo Liu and Ruoyu Chen and Chunwei Liu and GuoLiang Li and Jihua Kang and Fan Wu},
      year={2026},
      eprint={2605.03596},
      archivePrefix={arXiv},
      primaryClass={cs.AI},
      url={https://arxiv.org/abs/2605.03596}
}

@misc{shepard2026automationbench,
      title={AutomationBench}, 
      author={Daniel Shepard and Robin Salimans},
      year={2026},
      eprint={2604.18934},
      archivePrefix={arXiv},
      primaryClass={cs.AI},
      url={https://arxiv.org/abs/2604.18934}, 
}

@inproceedings{cheng2024seeclick,
    title = "{S}ee{C}lick: Harnessing {GUI} Grounding for Advanced Visual {GUI} Agents",
    author = "Cheng, Kanzhi  and
      Sun, Qiushi  and
      Chu, Yougang  and
      Xu, Fangzhi  and
      YanTao, Li  and
      Zhang, Jianbing  and
      Wu, Zhiyong",
    booktitle = "Proceedings of the 62nd Annual Meeting of the Association for Computational Linguistics (Volume 1: Long Papers)",
    month = aug,
    year = "2024",
    address = "Bangkok, Thailand",
    publisher = "Association for Computational Linguistics",
    url = "https://aclanthology.org/2024.acl-long.505",
    pages = "9313--9332"
}

@misc{rawles2023androidwildlargescaledataset,
      title={Android in the Wild: A Large-Scale Dataset for Android Device Control}, 
      author={Christopher Rawles and Alice Li and Daniel Rodriguez and Oriana Riva and Timothy Lillicrap},
      year={2023},
      eprint={2307.10088},
      archivePrefix={arXiv},
      primaryClass={cs.LG},
      url={https://arxiv.org/abs/2307.10088}, 
}

@article{wu2026mobilebench,
  title={MobileBench-OL: A Comprehensive Chinese Benchmark for Evaluating Mobile GUI Agents in Real-World Environment},
  author={Wu, Qinzhuo and Yang, Zhizhuo and Li, Hanhao and Gao, Pengzhi and Liu, Wei and Luan, Jian},
  journal={arXiv preprint arXiv:2601.20335},
  year={2026}
}

@misc{nie2026pspabenchpersonalizedbenchmarksmartphone,
      title={PSPA-Bench: A Personalized Benchmark for Smartphone GUI Agent}, 
      author={Hongyi Nie and Xunyuan Liu and Yudong Bai and Yaqing Wang and Yang Liu and Quanming Yao and Zhen Wang},
      year={2026},
      eprint={2603.29318},
      archivePrefix={arXiv},
      primaryClass={cs.AI},
      url={https://arxiv.org/abs/2603.29318}, 
}

@misc{sui2026androiddailyverifiablebenchmarkmobile,
      title={AndroidDaily: A Verifiable Benchmark for Mobile GUI Agents on Real-World Closed-Source Applications}, 
      author={Yifan Sui and Xin Huang and Hongbing Li and Fang Xu and Jiahe Lv and Haolong Yan and Yeqing Shen and Litao Liu and Zhimin Fan and Ziyang Meng and Jia Wang and Junbo Qi and Kaijun Tan and Zheng Ge and Xiangyu Zhang and Daxin Jiang and Osamu Yoshie},
      year={2026},
      eprint={2605.27761},
      archivePrefix={arXiv},
      primaryClass={cs.CV},
      url={https://arxiv.org/abs/2605.27761}, 
}

@misc{wu2026mobilegymverifiablehighlyparallel,
      title={MobileGym: A Verifiable and Highly Parallel Simulation Platform for Mobile GUI Agent Research},
      author={Dingbang Wu and Rui Hao and Haiyang Wang and Shuzhe Wu and Han Xiao and Zhenghong Li and Bojiang Zhou and Zheng Ju and Zichen Liu and Lue Fan and Zhaoxiang Zhang},
      year={2026},
      eprint={2605.26114},
      archivePrefix={arXiv},
      primaryClass={cs.AI},
      url={https://arxiv.org/abs/2605.26114}
}

@misc{jang2026iosworldbenchmarkpersonallyintelligent,
      title={iOSWorld: A Benchmark for Personally Intelligent Phone Agents}, 
      author={Lawrence Keunho Jang and Mareks Woodside and Geronimo Carom and Andrew Keunwoo Jang and Jing Yu Koh and Ruslan Salakhutdinov},
      year={2026},
      eprint={2606.09764},
      archivePrefix={arXiv},
      primaryClass={cs.LG},
      url={https://arxiv.org/abs/2606.09764}, 
}

@misc{anthropic2025buildingagents,
  author       = {Anthropic},
  title        = {Building Agents with the Claude Agent SDK},
  year         = {2025},
  month        = sep,
  publisher    = {Anthropic},
  howpublished = {\url{https://claude.com/blog/building-agents-with-the-claude-agent-sdk}},
}

@misc{anthropic2025effectiveharnesses,
  author       = {Anthropic},
  title        = {Effective Harnesses for Long-Running Agents},
  year         = {2025},
  month        = nov,
  publisher    = {Anthropic},
  howpublished = {\url{https://www.anthropic.com/engineering/effective-harnesses-for-long-running-agents}},
}

@article{yao2022react,
  title={ReAct: Synergizing Reasoning and Acting in Language Models},
  author={Yao, Shunyu and Zhao, Jeffrey and Yu, Dian and Du, Nan and Shafran, Izhak and Narasimhan, Karthik and Cao, Yuan},
  journal={arXiv preprint arXiv:2210.03629},
  year={2022}
}

@misc{shinn2023reflexionlanguageagentsverbal,
      title={Reflexion: Language Agents with Verbal Reinforcement Learning}, 
      author={Noah Shinn and Federico Cassano and Edward Berman and Ashwin Gopinath and Karthik Narasimhan and Shunyu Yao},
      year={2023},
      eprint={2303.11366},
      archivePrefix={arXiv},
      primaryClass={cs.AI},
      url={https://arxiv.org/abs/2303.11366}, 
}

@inproceedings{yang2024sweagent,
  title={{SWE}-agent: Agent-Computer Interfaces Enable Automated Software Engineering},
  author={John Yang and Carlos E Jimenez and Alexander Wettig and Kilian Lieret and Shunyu Yao and Karthik R Narasimhan and Ofir Press},
  booktitle={The Thirty-eighth Annual Conference on Neural Information Processing Systems},
  year={2024},
  url={https://arxiv.org/abs/2405.15793}
}

@inproceedings{wang2025openhands,
  title     = {OpenHands: An Open Platform for {AI} Software Developers as Generalist Agents},
  author    = {Xingyao Wang and Boxuan Li and Yufan Song and Frank F. Xu and Xiangru Tang and Mingchen Zhuge and Jiayi Pan and Yueqi Song and Bowen Li and Jaskirat Singh and Hoang H. Tran and Fuqiang Li and Ren Ma and Mingzhang Zheng and Bill Qian and Yanjun Shao and Niklas Muennighoff and Yizhe Zhang and Binyuan Hui and Junyang Lin and Robert Brennan and Hao Peng and Heng Ji and Graham Neubig},
  booktitle = {The Thirteenth International Conference on Learning Representations},
  year      = {2025},
  url       = {https://openreview.net/forum?id=OJd3ayDDoF}
}

@misc{openai2026harnessengineering,
  author       = {OpenAI},
  title        = {Harness Engineering: Leveraging {Codex} in an Agent-First World},
  year         = {2026},
  month        = feb,
  day          = {11},
  publisher    = {OpenAI},
  howpublished = {\url{https://openai.com/index/harness-engineering/}},
}

@misc{lin2026agenticharnessengineeringobservabilitydriven,
      title={Agentic Harness Engineering: Observability-Driven Automatic Evolution of Coding-Agent Harnesses}, 
      author={Jiahang Lin and Shichun Liu and Chengjun Pan and Lizhi Lin and Shihan Dou and Zhiheng Xi and Xuanjing Huang and Hang Yan and Zhenhua Han and Tao Gui and Yu-Gang Jiang},
      year={2026},
      eprint={2604.25850},
      archivePrefix={arXiv},
      primaryClass={cs.CL},
      url={https://arxiv.org/abs/2604.25850}, 
}

@misc{harnessbench2026,
  title     = {Harness Bench: Measuring Harness Effects in Realistic Agent Workflows},
  author    = {Harness Bench Team},
  year      = {2026},
  url       = {https://arxiv.org/abs/2605.27922},
  note      = {106 sandboxed offline agent tasks across 8 categories}
}

@misc{zhou2026hiconagenthistorycontextawarepolicy,
      title={HiconAgent: History Context-aware Policy Optimization for GUI Agents}, 
      author={Xurui Zhou and Gongwei Chen and Yuquan Xie and Zaijing Li and Kaiwen Zhou and Shuai Wang and Shuo Yang and Zhuotao Tian and Rui Shao},
      year={2026},
      eprint={2512.01763},
      archivePrefix={arXiv},
      primaryClass={cs.CV},
      url={https://arxiv.org/abs/2512.01763}, 
}

@article{xu2026mobile,
  title={Mobile-Agent-v3. 5: Multi-platform Fundamental GUI Agents},
  author={Xu, Haiyang and Zhang, Xi and Liu, Haowei and Wang, Junyang and Zhu, Zhaozai and Zhou, Shengjie and Hu, Xuhao and Gao, Feiyu and Cao, Junjie and Wang, Zihua and others},
  journal={arXiv preprint arXiv:2602.16855},
  year={2026}
}

@misc{li2026phoneharnessharnessingphoneuseagents,
      title={PhoneHarness: Harnessing Phone-Use Agents through Mixed GUI, CLI, and Tool Actions}, 
      author={Chenxin Li and Zhengyao Fang and Zhengyang Tang and Pengyuan Lyu and Xingran Zhou and Xin Lai and Fei Tang and Liang Wu and Yiduo Guo and Weinong Wang and Junyi Li and Yi Zhang and Yang Ding and Huawen Shen and Sunqi Fan and Shangpin Peng and Zheng Ruan and Anran Zhang and Benyou Wang and Chengquan Zhang and Han Hu},
      year={2026},
      eprint={2606.14832},
      archivePrefix={arXiv},
      primaryClass={cs.CL},
      url={https://arxiv.org/abs/2606.14832}, 
}

@misc{zheng2026taskstaterepresentationlonghorizonmobile,
      title={A Task-State Representation for Long-Horizon Mobile GUI Agents}, 
      author={Yujie Zheng and Zikang Liu and Xin Zhao and Ji-Rong Wen},
      year={2026},
      eprint={2607.00502},
      archivePrefix={arXiv},
      primaryClass={cs.CL},
      url={https://arxiv.org/abs/2607.00502}, 
}

@misc{qwen37plus,
    title = {{Qwen3.7-Plus}: Multimodal Agent Intelligence},
    url = {https://qwen.ai/blog?id=qwen3.7-plus},
    author = {{Qwen Team}},
    month = {May},
    year = {2026}
}

@misc{bytedance2026seed21,
  author       = {{ByteDance Seed}},
  title        = {{Seed2.1} {O}fficially {R}eleased: {A}dvancing {AI} {P}roductivity},
  year         = {2026},
  month        = jun,
  howpublished = {\url{https://seed.bytedance.com/en/blog/seed2-1-officially-released-advancing-ai-productivity}},
}

@misc{qwen38,
    title = {{Qwen3.8}: A New Bar for Coding and Cowork},
    url = {https://qwen.ai/blog?id=qwen3.8},
    author = {{Qwen Team}},
    month = {August},
    year = {2026}
}

@misc{googledeepmind2026gemini35flash,
  author       = {{Google DeepMind}},
  title        = {{Gemini 3.5 Flash} {M}odel {C}ard},
  year         = {2026},
  month        = may,
  day          = {19},
  howpublished = {\url{https://storage.googleapis.com/deepmind-media/Model-Cards/Gemini-3-5-Flash-Model-Card.pdf}},
}

@misc{openai2026gpt55,
  author       = {{OpenAI}},
  title        = {{GPT-5.5} {S}ystem {C}ard},
  year         = {2026},
  month        = apr,
  day          = {23},
  howpublished = {\url{https://deploymentsafety.openai.com/gpt-5-5/gpt-5-5.pdf}},
}

@misc{openai2026gpt56,
  author       = {{OpenAI}},
  title        = {{GPT-5.6} {S}ystem {C}ard},
  year         = {2026},
  month        = jul,
  day          = {9},
  howpublished = {\url{https://deploymentsafety.openai.com/gpt-5-6/gpt-5-6.pdf}},
}

@misc{anthropic2026claudesonnet46,
  author       = {{Anthropic}},
  title        = {{S}ystem {C}ard: {Claude Sonnet 4.6}},
  year         = {2026},
  month        = feb,
  day          = {17},
  howpublished = {\url{https://www-cdn.anthropic.com/bbd8ef16d70b7a1665f14f306ee88b53f686aa75/Claude%20Sonnet%204.6%20System%20Card.pdf}},
}

@misc{anthropic2026claudeopus47,
  author       = {{Anthropic}},
  title        = {{S}ystem {C}ard: {Claude Opus 4.7}},
  year         = {2026},
  month        = apr,
  day          = {16},
  howpublished = {\url{https://www-cdn.anthropic.com/037f06850df7fbe871e206dad004c3db5fd50340/Claude%20Opus%204.7%20System%20Card.pdf}},
}
\bibliographystyle{iclr2025_conference}


\end{document}